\documentclass[10pt, a4paper]{article}
\usepackage[hyphens]{url}
\usepackage{lrec2022} 
\usepackage{multibib}
\newcites{languageresource}{Language Resources}
\usepackage{graphicx}
\usepackage{tabularx}
\usepackage{soul}

\usepackage{url}
\usepackage{adjustbox}
\usepackage{caption}
\usepackage{expex}
\usepackage{enumitem}
\usepackage{multirow}
\usepackage{tikz}
\usepackage{wrapfig}
\usepackage{booktabs}
\usepackage{subcaption}
\usepackage{amsmath,amssymb}
\usepackage{dirtytalk}

\usepackage{todonotes}
\usepackage{xcolor}
\usepackage{microtype}

\newcommand{\unimorph}{UniMorph}
\newcommand{\feature}[1]{\textsc{\MakeLowercase{#1}}}

\newcommand{\citet}{\newcite}
\newcommand{\citep}{\cite}

\usepackage{titlesec}
\titleformat{\section}{\normalfont\large\bfseries\center}{\thesection.}{1em}{}
\titleformat{\subsection}{\normalfont\SmallTitleFont\bfseries\raggedright}{\thesubsection.}{1em}{}
\titleformat{\subsubsection}{\normalfont\normalsize\bfseries\raggedright}{\thesubsubsection.}{1em}{}
\renewcommand\thesection{\arabic{section}}
\renewcommand\thesubsection{\thesection.\arabic{subsection}}
\renewcommand\thesubsubsection{\thesubsection.\arabic{subsubsection}}

\usepackage{epstopdf}

\usepackage{hyperref}  \hypersetup{hidelinks}
\usepackage{xstring}

\usepackage{color}

\usepackage[T2A,T5,T1]{fontenc}
\usepackage[utf8]{inputenc}
\usepackage[russian, english]{babel}
\newcommand{\rus}[1]{\foreignlanguage{russian}{#1}}
\usepackage{arydshln}
\usepackage{mxedruli}
\usepackage{cjhebrew}
\newcommand{\geo}[1]{\begin{mxedr}#1\end{mxedr}}

\usepackage{tipa}
\newcommand{\jhu}{\textrm{\normalfont \textipa{Z}}}
\newcommand{\cmu}{\textrm{\normalfont \textbeltl}}
\newcommand{\num}{\textrm{\normalfont \textturnm}}
\newcommand{\unitn}{\textrm{\normalfont \textgamma}}
\newcommand{\ulou}{\textrm{\normalfont \textipa{3}}}
\newcommand{\york}{\textrm{\normalfont \textipa{7}}}
\newcommand{\ucam}{\textrm{\normalfont \textipa{Q}}}
\newcommand{\ethz}{\textrm{\normalfont \textipa{D}}}
\newcommand{\kul}{\textrm{\normalfont \textipa{V}}}
\newcommand{\umelb}{\textrm{\normalfont \textipa{@}}}
\newcommand{\hse}{\textrm{\normalfont \textipa{E}}}
\newcommand{\cub}{\textrm{\normalfont \textipa{X}}}
\newcommand{\krc}{\textrm{\normalfont \textipa{K}}}
\newcommand{\ubc}{\textrm{\normalfont \texthtb}}
\newcommand{\indu}{\textrm{\normalfont \textipa{P}}}
\newcommand{\uliv}{\textrm{\normalfont \textipa{L}}}
\newcommand{\nyu}{\textrm{\normalfont \textipa{5}}}
\newcommand{\swtb}{\textrm{\normalfont \textcloserevepsilon}}
\newcommand{\brian}{\textrm{\normalfont \textipa{B}}}
\newcommand{\nyuad}{\textrm{\normalfont \textcrh}}
\newcommand{\anu}{\textrm{\normalfont \textipa{S}}}
\newcommand{\biu}{\textrm{\normalfont \textlambda}}
\newcommand{\cdu}{\textrm{\normalfont \textipa{R}}}
\newcommand{\ugron}{\textrm{\normalfont \textscg}}
\newcommand{\ued}{\textrm{\normalfont \textsce}}
\newcommand{\pucp}{\textrm{\normalfont \textipa{H}}}
\newcommand{\nopoki}{\textrm{\normalfont \textipa{J}}}
\newcommand{\ugoth}{\textrm{\normalfont \texthtg}}
\newcommand{\sbras}{\textrm{\normalfont \textrtailz}}
\newcommand{\ilingran}{\textrm{\normalfont \texthardsign}}
\newcommand{\ispran}{\textrm{\normalfont \textscoelig}}
\newcommand{\helsinki}{\textrm{\normalfont \textbari}}
\newcommand{\tsu}{\textrm{\normalfont \textrtails}}
\newcommand{\neiu}{\textrm{\normalfont \textltailn}}
\newcommand{\bc}{\textrm{\normalfont \textctt}}
\newcommand{\swarthmore}{\textrm{\normalfont \textipa{F}}}
\newcommand{\uzh}{\textrm{\normalfont \ae}}
\newcommand{\ipipan}{\textrm{\normalfont \textipa{T}}}
\newcommand{\uindonesia}{\textrm{\normalfont \textthorn}}
\newcommand{\dbrau}{\textrm {\normalfont \textipa{U}}}
\newcommand{\mila}{\textrm{\normalfont \o}}
\newcommand{\msu}{\textrm{\normalfont \textipa{M}}}
\newcommand{\ddu}{\textrm{\normalfont \textipa{2}}}
\newcommand{\uore}{\textrm{\normalfont \textipa{9}}}
\newcommand{\buffalo}{\textrm{\normalfont \textomega}}

\newcommand{\lucid}{\textrm{\normalfont \textipa{C}}}
\newcommand{\bgu}{\textrm{\normalfont \textipa{G}}}
\newcommand{\sbu}{\textrm{\normalfont \textipa{Y}}}
\newcommand{\cuny}{\textrm{\normalfont \textipa{N}}}
\newcommand{\inria}{\textrm{\normalfont \textltailn }}
\newcommand{\georgetown}{\textrm{\normalfont \texthtrtaild}}
\newcommand{\gmu}{\textrm{\normalfont \textipa{\!j}}}
\newcommand{\athena}{\textrm{\normalfont \textipa{\!G}}}
\newcommand{\printfnsymbol}[1]{%
  \textsuperscript{\@fnsymbol{#1}}%
}

\title{UniMorph 4.0: Universal Morphology}
\name{Khuyagbaatar Batsuren$^\num$\sthanks{~~The authors contributed equally}~\;~Omer Goldman$^\biu$\footnotemark[1]~~\;~Salam Khalifa$^\sbu$~\;~Nizar Habash$^\nyuad$\\ 
{\bf \large Witold Kieraś$^\ipipan$~\;~Gábor Bella$^\unitn$~\;~Brian Leonard$^\brian$~\;~Garrett Nicolai$^\ubc$~\;~Kyle Gorman$^\cuny$}\\
{\bf \large Yustinus Ghanggo Ate$^\swtb$~\;~Maria Ryskina$^\cmu$~\;~Sabrina Mielke$^\jhu$~\;~Elena Budianskaya$^\ilingran$}\\
{\bf \large Charbel El-Khaissi$^\anu$~\;~Tiago Pimentel$^\ucam$~\;~Michael Gasser$^\indu$~\;~William Lane$^\cdu$~\;~Mohit Raj$^\dbrau$}\\
{\bf \large Matt Coler$^\ugron$~\;~Jaime Rafael Montoya Samame$^\pucp$~\;~Delio Siticonatzi Camaiteri$^\nopoki$~\;~Beno\^it Sagot$^\inria$}\\
{\bf \large Esaú Zumaeta Rojas$^\nopoki$~\;~Didier López Francis$^\nopoki$~\;~Arturo Oncevay$^\ued$~\;~Juan López Bautista$^\nopoki$} \\
{\bf \large Gema Celeste Silva Villegas$^\pucp$~\;~Lucas Torroba Hennigen$^\ucam$~\;~Adam Ek$^\ugoth$~David Guriel$^\biu$}\\
{\bf \large 
Peter~Dirix$^\kul$~\;~Jean-Philippe Bernardy$^\ugoth$~\;~Andrey Scherbakov$^\umelb$~\;~Aziyana Bayyr-ool$^\sbras$}\\
{\bf \large Antonios Anastasopoulos$^\gmu$~\;~Roberto Zariquiey$^\pucp$~\;~Karina Sheifer$^{\hse,\ilingran,\ispran}$~\;~Sofya Ganieva$^{\msu,\ilingran}$} \\ 
{\bf \large Hilaria Cruz$^\ulou$~\;~Ritván Karahóǧa$^\athena$~\;~Stella Markantonatou$^\athena$~\;~George Pavlidis$^\athena$} \\ 
{\bf \large Matvey Plugaryov$^{\msu,\ilingran}$~\;~Elena Klyachko$^{\hse,\ilingran}$~\;~Ali Salehi$^\buffalo$~\;~Candy Angulo$^\pucp$~\;~Jatayu Baxi$^\ddu$} \\
{\bf \large Andrew Krizhanovsky$^\krc$~\;~Natalia Krizhanovsky$^\krc$~\;~Elizabeth Salesky$^\jhu$~\;~Clara Vania$^\nyu$} \\
{\bf \large Sardana Ivanova$^\helsinki$~\;~Jennifer White$^\ucam$~\;~Rowan Hall Maudslay$^\ucam$~\;~Josef Valvoda$^\ucam$}\\
{\bf \large Ran Zmigrod$^\ucam$~\;~Paula Czarnowska$^\ucam$~\;~Irene Nikkarinen$^\ucam$~\;~Aelita Salchak$^\tsu$~\;~Brijesh Bhatt$^\ddu$}\\
{\bf \large Christopher Straughn$^\neiu$~\;~Zoey Liu$^\bc$~\;~Jonathan North Washington$^\swarthmore$~\;~Yuval Pinter$^\bgu$} \\
{\bf \large Duygu Ataman$^\nyu$~\;~Marcin Woliński$^\ipipan$~\;~Totok Suhardijanto$^\uindonesia$~\;~Anna Yablonskaya$^\hse$} \\
{\bf \large Niklas Stoehr$^\ethz$~\;~Hossep Dolatian$^\sbu$~\;~Zahroh Nuriah$^\uindonesia$~\;~Shyam Ratan$^\dbrau$~\;~Francis M. Tyers$^{\indu,\hse}$} \\
{\bf \large Edoardo M. Ponti$^\mila$~\;~Grant Aiton$^\anu$~\;~Aryaman Arora$^\georgetown$~\;~Richard J. Hatcher$^\buffalo$} \\
{\bf \large Ritesh Kumar$^\dbrau$~\;~Jeremiah Young$^\uore$~\;~Daria Rodionova$^\hse$~\;~Anastasia Yemelina$^\hse$}\\
{\bf \large Taras Andrushko$^\hse$~\;~Igor Marchenko$^\hse$~\;~Polina Mashkovtseva$^\hse$~\;~Alexandra Serova$^\hse$} \\
{\bf \large Emily Prud'hommeaux$^\bc$~\;~Maria Nepomniashchaya$^\hse$~\;~Fausto Giunchiglia$^\unitn$} \\
{\bf \large Eleanor Chodroff$^\york$~\;~Mans Hulden$^\cub$~\;~Miikka Silfverberg$^\ubc$~\;~Arya D. McCarthy$^\jhu$}\\
{\bf \large David Yarowsky$^\jhu$~\;~Ryan Cotterell$^\ethz$~\;~Reut Tsarfaty$^\biu$~\;~Ekaterina Vylomova$^\umelb$}
}

\address{$^\num$National University of Mongolia $^\biu$Bar-Ilan University $^\jhu$Johns Hopkins University $^\unitn$University of Trento \\
$^\york$University of York $^\cmu$Carnegie Mellon University $^\brian$Brian Leonard Consulting $^\indu$Indiana University\\
$^\ubc$University of British Columbia $^\ddu$Dharmsinh Desai University $^\nyuad$New York University Abu Dhabi  $^\inria$Inria\\ $^\ucam$University of Cambridge $^\ugoth$University of Gothenburg $^\uore$University of Oregon $^\anu$Australian National University\\
$^\athena$ILSP/Athena RC $^\ugron$University of Groningen $^\kul$KU Leuven $^\ulou$University of Louisville $^\ued$University of Edinburgh \\
$^\pucp$Pontificia Universidad Católica del Perú 
$^\nopoki$Universidad Católica Sedes Sapientiae, Filial Atalaya\\ 
$^\sbras$Institute of Philology of the Siberian Branch of the Russian Academy of Sciences $^\msu$Moscow State University \\
$^\bc$Boston College $^\hse$Higher School of Economics $^\ilingran$Institute of Linguistics, Russian Academy of Sciences \\
$^\uzh$University of Z{\"u}rich $^\swtb$STKIP Weetebula $^\ispran$Institute for System Programming, Russian Academy of Sciences \\
$^\buffalo$University at Buffalo $^\krc$Karelian Research Centre of the Russian Academy of Sciences $^\swarthmore$Swarthmore College\\
$^\lucid$ESRC International Centre for Language and Communicative Development(LuCiD) $^\nyu$New York University \\ $^\neiu$Northeastern Illinois University $^\helsinki$University of Helsinki $^\tsu$Tuvan State University $^\georgetown$Georgetown University \\ $^\cdu$Charles Darwin University $^\ipipan$Institute of Computer Science, Polish Academy of Sciences $^\uindonesia$Universitas Indonesia \\ $^\sbu$Stony Brook University $^\dbrau$Dr.~Bhimrao Ambedkar University
$^\mila$Mila/McGill University Montreal \\ $^\cub$University of Colorado Boulder $^\uliv$University of Liverpool $^\cuny$Graduate Center, City University of New York\\ $^\gmu$George Mason University $^\bgu$Ben-Gurion University of the Negev $^\ethz$ETH Z{\"u}rich $^\umelb$University of Melbourne\\
\texttt{khuyagbaatar@num.edu.mn}\quad \texttt{omer.goldman@gmail.com}\quad \texttt{vylomovae@unimelb.edu.au}}

\abstract{
The Universal Morphology (UniMorph) project is a collaborative effort providing broad-coverage instantiated normalized morphological inflection tables for hundreds of diverse world languages.
The project comprises two major thrusts: a language-independent feature schema for rich morphological annotation and a type-level resource of annotated data in diverse languages realizing that schema.
This paper presents the expansions and improvements made on several fronts over the last couple of years (since \newcite{mccarthy-etal-2020-unimorph}). Collaborative efforts by numerous linguists have added 67 
new languages, including 30 endangered languages.
We have implemented several improvements to the extraction pipeline to tackle some issues, e.g.\ missing gender and macron information. 
We have also amended the schema to use a hierarchical structure that is needed for morphological phenomena like multiple-argument agreement and case stacking, while adding some missing morphological features to make the schema more inclusive.
In light of the last UniMorph release, we also augmented the database with morpheme segmentation for 16 languages.
Lastly, this new release makes a push towards inclusion of derivational morphology in UniMorph by enriching the data and annotation schema with instances representing derivational processes from MorphyNet.
}

\usepackage{gb4e} \noautomath
\begin{document}

\maketitleabstract

\section{Introduction}

Developing categories that allow for cross-linguistic comparison is one of the most challenging tasks in linguistic typology.
Typologists have proposed dimensions of cross-linguistic variation such as fusion \cite{wals-20}, inflectional synthesis \cite{wals-22}, position of case affixes \cite{wals-51}, number of cases \cite{wals-49}, and others,
and these dimensions and descriptions are being progressively refined~\cite{haspelmath2007pre}.

\citet{evans2009myth} critically discuss the idea of ``linguistic universals'', demonstrating extensive diversity across all levels of linguistic organization.
 The distinction between \textit{g-linguistics}, a study of Human Language in general, and \textit{p-linguistics}, a study of particular languages, including their idiosyncratic properties, is discussed in \citet{haspelmath2021general}.
The UniMorph annotation schema~\cite{sylak-glassman-etal-2015-language}, and this work in particular, is an attempt to balance the trade-off between descriptive categories and comparative concepts through a more fine-grained analysis of languages \citep{haspelmath2010comparative}. 
The initial schema~\cite{10.1007/978-3-319-23980-4_5} was based on the analysis of typological literature and included 23 dimensions of meaning (such as tense, aspect, grammatical person, number) and over 212 features (such as past/present for tense or singular/plural for number). 
The first release of the UniMorph database included 8 languages extracted from the English edition of Wiktionary~\cite{kirov2016very,cotterell-etal-2016-sigmorphon}.
The database has been augmented with 52 and 66 new languages in versions 2.0 and 3.0, respectively~\cite{kirov-etal-2018-unimorph,mccarthy-etal-2020-unimorph}.
UniMorph 3.0 introduced many under-resourced languages derived from various linguistic sources.
Prior to each release, all language datasets were included in part in the SIGMORPHON shared tasks on morphological reinflection \cite{cotterell-etal-2016-sigmorphon,cotterell-etal-2017-conll,cotterell-etal-2018-conll,mccarthy-etal-2019-sigmorphon}. The current release includes languages of the 2020--2021 shared tasks~\cite{vylomova-etal-2020-sigmorphon,pimentel2021sigmorphon}.
Unlike previous versions, linguistic data comes from grammar descriptions and finite-state models.  

\begin{figure}[t]
\centering
\includegraphics[width=1\linewidth]{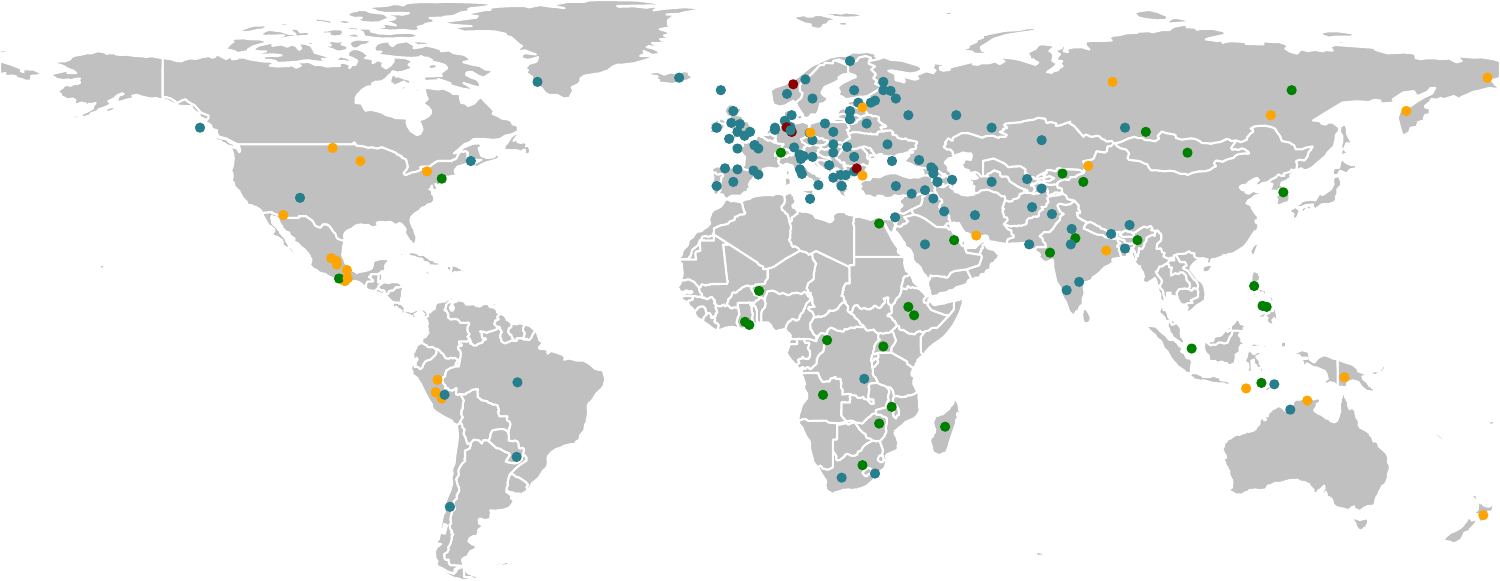}
\caption{The UniMorph 4.0 languages (Oranges are endangered, dark reds are historic, greens are new languages, and blues are old languages).}
\label{fig:map}
\end{figure}

The work described here, representing the UniMorph 4.0 milestone, makes several contributions to further improve the UniMorph data and tools.
First, we include inflection tables for 67 new languages and extend the datasets for 
31 languages, increasing the total number of languages to 182.
We note that the upcoming decade 2022--2032 has been announced as the Decade on Indigenous Languages,\footnote{\url{https:///en.unesco.org/news/upcoming-decade-indigenous-languages-2022-2032-focus-indigenous-language-users-human-rights}} and in this release we are enriching the UniMorph database with 30 endangered languages, as listed by UNESCO.\footnote{\url{http://www.unesco.org/languages-atlas/index.php}} 
Second, we update the annotation schema to improve representation of phenomena such as polypersonal agreement and case stacking.
Third, we provide morpheme segmentation data for 16 languages. 
Fourth, we introduce morpheme-annotated dataset of derivational morphology in 30 languages.
Finally, we release new automatic validation tool to evaluate UniMorph against Universal Dependencies treebanks \citep{nivre2016universal}.
On the whole, UniMorph 4.0 covers 182 languages (as shown in Figure \ref{fig:map}), 122M inflections, and 769K derivations.

\section{Schema Updates} \label{sec:schema}

\subsection{Hierarchical Annotation}

\begin{table*}[ht]
\centering
\small{
\begin{tabular}{@{}llll@{}}  \toprule
Language & Form & Hierarchical Schema & Flat Schema \\ \midrule
English & drinks & \textsc{v;prs;nom(3,sg)} & \textsc{v;prs;3;sg} \\
Georgian & \geo{gagi+svebt} & \textsc{v;fut;nom(1,pl);acc(2,sg)} & \textsc{v;fut;argno1p;argac2s} \\
Hebrew & \cjRL{`mdth} & \textsc{n;sg;pssd;pss(3,sg,fem)} & \textsc{n;sg;pssd;pss3sf} \\ 
Russian & \rus{собакам} & \textsc{n;dat(pl)} & \textsc{n;dat;pl} \\
Evenki & \rus{ңинакиннундуле} & \textsc{n;all(com(sg))} & --- \\ 
Turkish & kedisini & \textsc{n;acc(sg;pssd;pss(1,sg))} & \textsc{n;sg;acc;pssd;pss1s}\\ \bottomrule
\end{tabular}
}
\caption{Example hierarchically annotated forms, including treatment of arguments, cases or both.}
\label{tab:hierarchy}
\end{table*}

The major structural change to the annotation schema in this release is the introduction of a hierarchical feature structure, following \newcite{guriel-etal-2022-morphological}, instead of the flat structure that characterized the schema thus far. The shift is done to allow smoother incorporation of data for some non-western languages while keeping it easy to process. Specifically, the hierarchy is needed to annotate case stacking, polypersonal agreement, and more---treatment of some of which is impossible under the current system.

Verb forms with polypersonal agreement agree with more than one argument of the verb. In contrast to most western languages, where the verb agrees only with the subject (in the nominative case), verbs in many languages may agree with up to four different arguments. The existing schema attributes nominative features directly to the verbs in languages where only nominative agreement exists. Thus, for example, the English form \textit{drinks} is annotated as \textsc{v;prs;3;sg}, where the nominative-related features \textsc{3;sg} are on the same level as \textsc{prs}. However, for languages with poly-personal agreement a case specification is needed, and the solution is to mark that in a composite feature like \textsc{argac1s} for a case where a form agrees with the verb’s accusative argument which is 1st person singular.

The updated schema 
places the treatment of both cases on equal ground, while unpacking the composite feature string to a decomposable feature structure.
Following \citet{anderson-1992-morphous}, features are \textit{layered} such that some features may be composed of another set of features from the same feature inventory. We employ this structure to annotate every argument as a complex feature that includes all features pertaining to that argument. The aforementioned feature \textsc{argac1s} is thus replaced with the composite feature \textsc{acc(1,sg)}, and a form that was formerly annotated as \textsc{v;prs;argno3p;argac2s} is annotated as \textsc{v;prs;nom(3,pl);acc(2,sg)}.
This solution applies not only to poly-personal agreement, but to any case in which annotation of a single form requires more than one person-number-gender feature bundle, like in the case of possessed nominals. See Table \ref{tab:hierarchy} for detailed examples.

Another case that requires hierarchical annotation is case stacking. In this phenomenon a noun takes the case suffix of its nominal head in addition to its own case suffix. For example in Evenki:
\begin{exe}
\ex
\glll \rus{асаткандула} \rus{ңинакиннундуле}\\
asatkan-dula nginakin-nun-dule \\
girl.\feature{ALL} dog.\feature{COM}.\feature{ALL} \\
\glt `to the girl with the dog'
\end{exe}
In these cases, the order of the cases is essential, but it cannot be captured by a flat unordered set of features. Therefore, in the updated schema cases are applied on top of the other nominal features and a form that was formerly tagged as \textsc{n;sg;nom} would now be tagged as \textsc{n;nom(sg)}.
This allows application of multiple cases in an order-preserving manner such that \textsc{n;all(com(sg))} is different from \textsc{n;com(all(sg))}.

For backward compatibility, the previous flat schema will continue to be maintained, although it cannot treat all forms in some extreme cases.

\subsection{Derivational Morphology}
UniMorph 4.0 releases a dataset of derivational morphology in 30 languages, annotated with morphemes and morphological features. The lemma (source word form) and derivation (target word form) are related to particular morphological annotation features represented by common part-of-speech tags and morpheme, as in the Italian example of \textit{morfologia} `morphology' and \textit{morfologico} `morphological':
\begin{center}
( \textit{morfologia},~~~\textit{morfologico},~~~N:ADJ,~~~`\textit{-ico}' ),
\end{center}
and in the French example of \textit{décrit} `to describe' and \textit{susdécrit} `above described':
\begin{center}
( \textit{décrit},~~~\textit{susdécrit},~~~V:ADJ,~~~`\textit{sus-}' ).
\end{center}
Compared to state-of-the-art derivational resources \citep{vidra-etal-2019-derinet,kyjanek-etal-2019-universal}, this dataset provides explicit morphemes between source and target word forms. With these morphemes, subword tokenization \citep{sennrich2016neural,mielke2021between}  can be advanced to dictionary-based morpheme segmentation for derivationally rich languages like English and French. The extraction process and results of the derivational dataset are presented in Section \ref{section:derivation}. 


\subsection{New Morphosyntactic Features}
\textbf{Mood.} The UniMorph schema \cite{10.1007/978-3-319-23980-4_5} combines imperative and jussive moods under one tag (\textsc{imp}). This creates inconsistencies for languages such as Arabic. In Modern Standard Arabic (MSA), a verb can be perfective, imperfective or imperative (often marked as their aspect). Perfective verbs are always indicative, imperative verbs don't usually express mood, and imperfective verbs can be indicative, subjunctive, or jussive. To be able to transparently describe verbs in MSA, we split the imperative--jussive tag into two tags: imperative (\textsc{imp}) and jussive (\textsc{jus}), to accommodate  imperative verbs and imperfective--jussive verbs.

\textbf{Argument Marking.} While working on indigenous languages of the Americas, Australia and Russia, we augmented the schema with the following features for argument marking: \textsc{no1}, \textsc{no2}, \textsc{no3}, \textsc{no3f}, \textsc{no3m}, \textsc{ac1},  \textsc{ac2},  \textsc{ac3}  (no number specified), \textsc{no1pi}, \textsc{no1pe} (adding inclusivity), \textsc{ac1d}, \textsc{ac2d}, \textsc{ac3d} (adding dual number).\footnote{Although the annotation guidelines dictate that all argument marking features have an \textsc{arg} prefix, in practice it is omitted for all argument features.}

\textbf{Possession.} We added the following tags: \textsc{pss1i} (1st person inclusive), \textsc{pss3f}, \textsc{pss3m} (gender-specific tags), \textsc{pssrs} and \textsc{pssrp} (reflexive singular and plural).


\begin{table*}[!h]
\begin{adjustbox}{width=1\textwidth}
\small
\begin{tabular}{@{}p{0.7in}p{0.7in}p{0.4in}p{1in}p{1in}p{2in}p{1in}@{}}
\toprule
 Family & Genus & ISO & Language & Source of Data & Annotators & Lemmas/Forms \\
 \midrule
 Afro-Asiatic  & Semitic &  \texttt{afb}  & Gulf Arabic  & \citet{khalifa-etal-2018-morphologically-2} & Salam Khalifa, Nizar Habash  &  6,345/24,077\\
   & Semitic &  \texttt{amh} & Amharic  & \citet{hornmorpho} & Michael Gasser   &  2,461/46,224\\
  & Semitic &  \texttt{arz}   & Egyptian Arabic  & \citet{habash-etal-2012-morphological} & Salam Khalifa, Nizar Habash  & 6,004/17,009 \\ 
  &  Cushitic & \texttt{orm} & Oromo & \citet{oromo} & Irene Nikkarinen & 92/2,046\\
 \midrule
 Algic & Algonquian & \texttt{cre$^*$} & Plains Cree & \citet{cree}  & Eleanor Chodroff & 32/9,577  \\
  \midrule
 Arawakan  & Southern Arawakan&  \texttt{ame$^*$}  & Yanesha'  & \citet{duff-1980-diccionario} & Gema Celeste Silva Villegas, Juan López Bautista, Didier López Francis, Roberto Zariquiey, Arturo Oncevay  & 327/3,767 \\
          & Southern Arawakan &  \texttt{cni$^*$}  & Asháninka  & \citet{zumaeta-zerdin-2018-guia,kindberg-1980-diccionario} & Jaime Rafael Montoya Samame, Esaú Zumaeta Rojas, Delio Siticonatzi C., Roberto Zariquiey, Arturo Oncevay  & 407/20,070\\ 
  \midrule
Austronesian     & Malayo-Polynesian &  \texttt{ind}  &  Indonesian  &  \href{https://kbbi.web.id/}{KBBI}, Wikipedia & Clara Vania, Totok Suhardijanto, Zahroh Nuriah &  3,877/27,714\\
 &  &  \texttt{kod$^*$}  &  Kodi  & \citet{GhanggoAte2021Doc} & Yustinus Ghanggo Ate, Garrett Nicolai & 64/463 \\
   \cline{2-7}
& \multirow{3}{*}{\shortstack[l]{Greater \\Central \\ Philippine}}   & \texttt{ceb}   &Cebuano  & \citet{cebuano}     & Ran Zmigrod    &    97/618  \\
    &   & \texttt{hil}  & Hiligaynon  & \citet{hiligaynon} &  Ran Zmigrod & 97/1,256\\
    &   & \texttt{tgl}   &Tagalog     & \citet{tagalog}    &  Jennifer White & 344/2,912 \\
    \cline{2-7}
    &Oceanic  & \texttt{mri$^*$}  & M\=aori      & \citet{maori}     &  Jennifer White   & 104/214    \\
    \cline{2-7}
    &Barito   &\texttt{mlg}   &Malagasy    &\citet{malagasy} & Jennifer White &  159/644\\
 \midrule
Aymaran  & Aymaran &  \texttt{aym}  & Aymara  & \citet{coler2014grammar} & Matt Coler, Eleanor Chodroff  & 3,410/336,341\\
 \midrule
Chukotko-Kamchatkan  & 	Northern Chukotko-Kamchatkan &  \texttt{ckt$^*$}  & Chukchi  &  \href{https://chuklang.ru/}{Chuklang};  \citet{tyers-mishchenkova-2020-dependency} & Karina Sheifer, Maria Ryskina  & 197/243\\
\cline{2-7}
 & Southern Chukotko-Kamchatkan &  \texttt{itl$^*$} & Itelmen  &  & Karina Sheifer, Sofya Ganieva, Matvey Plugaryov &  1,636/2,701\\
 \midrule
Gunwinyguan & Gunwinggic &  \texttt{gup$^*$}  & Kunwinjku  &  \citet{lane-bird-2019-towards-2} & William Lane  &  73/307\\
  \midrule
\multirow[t]{11}{*}{Indo-European}  &  Indic &  \texttt{asm}  &  Assamese  & Wiktionary & Khuyagbaatar Batsuren, Aryaman Arora  & 1,877/94,147\\
   &   &  \texttt{bra}  &  Braj  & \citet{Kumar:2018} & Shyam Ratan, Ritesh Kumar  & 1,246/1,821\\
   &  &  \texttt{mag$^*$}  &  Magahi  & \citet{kumar_developing_2014}  & Mohit Raj, Ritesh Kumar  & 1,612/2,194\\
   &   &  \texttt{guj}  &  Gujarati  & \citet{baxi2021morpheme};Wiktionary & Jatayu Baxi, Brijesh S. Bhatt, Khuyagbaatar Batsuren, Aryaman Arora  & 6,995/19,404 \\
   &   &  \texttt{hsi$^*$}  &  Kholosi  & \citet{arora-2021-kholosi-dictionary} & Aryaman Arora  & 49/174\\
   \cline{2-7}
  & Germanic & \texttt{afr} & Afrikaans & \citet{afritools} & Peter Dirix  &  179,941/309,558\\
  &  & \texttt{gsw} & Swiss German & \citet{ch} & Ryan Cotterell  &  145/2067\\
  & & \texttt{got} & Gothic & Wiktionary & Khuyagbaatar Batsuren (KB)  & 4,126/102,083 \\
  & & \texttt{goh} & Old High German & Wiktionary & Jeremiah Young; KB  & 482/7,248 \\
  & & \texttt{non} & Old Norse & Wiktionary & Jeremiah Young; KB & 2,520/98,185 \\
  \cline{2-7}
  & Slavic & \texttt{slk} & Slovak & \citet{morfflexsk} & Witold Kieraś & 366,183/28,428,612 \\
  &  & \texttt{hsb$^*$} & Upper Sorbian & \citet{fraser-2020-findings} & Taras Andrushko, Igor Marchenko & 310/400 \\
  &  & \texttt{poma} & Pomak & under review  & Ritván Karahóǧa, Stella Markantonatou, Georgios Pavlidis, Antonios Anastasopouos & 233,533/6,557,759\\
   \midrule
Iroquoian  & Northern Iroquoian &  \texttt{see$^*$}  & Seneca  & ~\citet{phyllis} & Richard J. Hatcher, Emily Prud'hommeaux, Zoey Liu  & 5,430/140\\
 \midrule
Koreanic  & Koreanic &  \texttt{kor}  & Korean  & Wiktionary & Maria Nepomniashchaya, Daria Rodionova, Anastasia Yemelina  & 2,686/241,323\\
 \midrule
Mongolic & Mongolic &  \texttt{khk}   & Khalkha Mongolian  & \citet{munkhjargal2016morphological,batsuren-etal-2019-building} & Khuyagbaatar Batsuren   &  2,085/14,592\\
 \midrule
Niger--Congo & Bantoid   & \texttt{kon}   &Kongo& \citet{kongo} & Jennifer White & 200/828 \\
    &    &\texttt{lin}   &Lingala & \citet{lingala} & --- & 57/228\\
    &   &\texttt{lug}   &Luganda     &\citet{luganda} & Edoardo M. Ponti & 89/4,895\\
    &    &\texttt{nya}   &Chewa  & \citet{chewa} & Ryan Cotterell  & 227/4,370\\
    & &\texttt{sot}   &Sotho & \citet{sotho}       &  --- &  26/494 \\
    &    &\texttt{sna}   & Shona &\citet{shona,shona2}   & Rowan Hall Maudslay  & 86/3,030 \\
    \cline{2-7}
    & Kwa   &\texttt{aka}   &Akan & \citet{akan}  & Tiago Pimentel & 96/4,182\\
    &   &\texttt{gaa}   &Gã       & \citet{ga}  & Tiago Pimentel & 95/909\\
\bottomrule
\end{tabular}
\end{adjustbox}
\caption{Inflectional paradigms: new languages (Endangered languages are marked with $^*$)} 
\label{tab:new-lang-descr}
\end{table*}

\begin{table*}[!h]
\begin{adjustbox}{width=1\textwidth}
\small
\begin{tabular}{@{}p{0.5in}p{0.7in}p{0.4in}p{1.2in}p{1.4in}p{1.6in}p{1in}@{}}
\toprule
 Family & Genus & ISO & Language & Source of Data & Annotators & Lemmas/Forms\\
 \midrule
Oto-Manguean    &Amuzgoan  &\texttt{azg$^*$}   &San Pedro Amuzgos Amuzgo      & \citet{azg2015}   &  Antonis Anastasopoulos  & 332/12,204\\
    &Chichimec  &\texttt{pei$^*$}   &Chichimeca-Jonaz      & \citet{pei2015} & Antonis Anastasopoulos    & 123/15,120\\
    &Chinantecan   &\texttt{cpa$^*$}  &Tlatepuzco Chinantec       & \citet{cpa2015} & Antonis Anastasopoulos        & 697/7,893\\
    &Mixtecan   &\texttt{xty}   &Yoloxóchitl Mixtec      & \citet{xty2015}  &      Antonis Anastasopoulos     & 594/3,057 \\
    &Otomian   &\texttt{ote$^*$}   &Mezquital Otomi  & \citet{ote2015}  &     Antonis Anastasopoulos       &  2,028/33,162\\
    &Otomian  &\texttt{otm$^*$}   &Sierra Otomi  & \citet{otm2015}   &  Antonis Anastasopoulos  & 1,909/31,380\\
    &Zapotecan    &\texttt{cly$^*$}   & Eastern Chatino of San Juan Quiahije &    \citet{cruz-anastasopoulos-stump:2020:LREC}  &   Hilaria Cruz, Antonis Anastasopoulos  & 185/4,716\\
    &Zapotecan &\texttt{ctp$^*$}   & Eastern Chatino of Yaitepec   & \citet{ctp2015} &  Antonis Anastasopoulos    & 223/3,796 \\
    &Zapotecan  &\texttt{czn$^*$}   &Zenzontepec Chatino  & \citet{czn2015}  &         Antonis Anastasopoulos &  386/1,567 \\
    &Zapotecan    &\texttt{zpv$^*$}   &Chichicapan Zapotec      & \citet{zpv2015}    &   Antonis Anastasopoulos & 379/1,164\\
    \midrule
Pano-Tacana & Pano & \texttt{shp$^*$} & Shipibo-Konibo & \citet{james1993diccionario};\citet{valenzuela2003transitivity} & Candy Angulo, Roberto Zariquiey, Arturo Oncevay & 2,111/14,588 \\
  \midrule
Siouan & Core Siouan & \texttt{dak$^*$} & Dakota & \citet{dakota} & Eleanor Chodroff & 537/3,766 \\
  \midrule
Songhay & Songhay & \texttt{dje} & Zarma & \citet{zarma} & Ran Zmigrod & 27/84\\
  \midrule
 Trans-New Guinea  & Bosavi &  \texttt{ail$^*$}  &  Eibela  &  \citet{aiton2016grammar} & Grant Aiton, Edoardo Maria Ponti, Ekaterina Vylomova  & 642/2,718\\
  \midrule
Tungusic  & Tungusic &  \texttt{evn$^*$}  & Evenki  &  \citet{kazakevich2013creating} & Elena Klyachko & 4,495/11,371\\

   & Tungusic &  \texttt{sjo$^*$}   & Xibe  &  \citet{zhou-etal-2020-universal} & Elena Klyachko   & 1,892/3,054\\
 \midrule
Turkic  & Turkic &  \texttt{sah}  & Sakha  &  \citet[Apertium: \href{https://github.com/apertium/apertium-sah}{\texttt{apertium-sah}}]{forcada2011apertium} & Francis M. Tyers, Jonathan North Washington, Sardana Ivanova, Christopher Straughn, Maria Ryskina  & 5,622/590,765\\
  & Turkic &  \texttt{tyv}  & Tuvan  &   \citet[Apertium: \href{https://github.com/apertium/apertium-tyv}{\texttt{apertium-tyv}}]{forcada2011apertium} & Francis M. Tyers, Jonathan North Washington, Aziyana Bayyr-ool, Aelita Salchak, Maria Ryskina  & 5,032/586,180 \\
 & Turkic &  \texttt{kir}  & Kyrgyz  & \cite{kyrgyz} & Eleanor Chodroff & 98/5,544\\
 & Turkic &  \texttt{uig}  & Uyghur  & \cite{uyghur} & Eleanor Chodroff  & 90/8,178 \\
 & Turkic &  \texttt{uzb}  & Uzbek  & \cite{uzbek,uzbek2} & Eleanor Chodroff & 428/36,031\\
 \midrule
 Uralic & Finnic &  \texttt{vro$^*$}   & Võro  & \citet{iva2007voru}&Ekaterina Vylomova   &  63/512\\
 \midrule
 Uto-Aztecan & Tepiman & \texttt{ood$^*$} & O’odham & \citet{oodham} & Eleanor Chodroff  & 370/1,628\\
 \midrule
Yeniseian   &   Northern Yeniseian &   \texttt{ket$^*$} &  Ket &  \href{https://github.com/lalsnivts/ket_corpus}{Ket corpus}  &  Elena Budianskaya, Polina Mashkovtseva, Alexandra Serova &    349/1,184\\
 \midrule
 Constructed & --- & \texttt{epo} & Esperanto & Wiktionary & Arya D. McCarthy & 1,945/58,350 \\
\bottomrule
\end{tabular}
\end{adjustbox}
\caption{Inflectional paradigms: new languages (continuation; Endangered languages are marked with $^*$) }
\label{tab:new-lang-descr-2}
\end{table*}

\begin{table*}[!h]
\begin{adjustbox}{width=1\textwidth}
\small
\begin{tabular}{@{}p{0.7in}p{0.7in}p{0.4in}p{1in}p{1.5in}p{1.5in}p{1in}@{}}
\toprule
 Family & Genus & ISO & Language & Source of Data & Annotators  & Lemmas/Forms\\
 \midrule
Afro-Asiatic   & Semitic &  \texttt{ara}  & Standard Arabic  & \citet{taji:2018:arabic} & Salam Khalifa, Nizar Habash  &  11,676/418,010\\ 
    & Semitic &  \texttt{heb}  & Hebrew (Vocalized)  & Wiktionary &Omer Goldman  & 1,183/33,178\\
  & Semitic &  \texttt{heb}  & Hebrew (Unvocalized)  &  \citet{sade2018hebrew} & Anna Yablonskaya  & 6,499/14,454\\
  & Semitic &  \texttt{syc}  & Classic Syriac  &  \href{https://sedra.bethmardutho.org/}{SEDRA} & Charbel El-Khaissi  & 3,299/31,972\\
  \midrule
 Indo-European & Iranian &  \texttt{ckb}  & Central Kurdish (Sorani)  &  Alexina project  & Ali Salehi  & 274/22,990\\
   & Iranian &  \texttt{sdh}  &  Southern Kurdish & \citet[native speakers]{fattah2000dialectes} & Ali Salehi & 1/189\\
   & Romance &  \texttt{fra}  & French  &  \citet{sagot-2010-lefff}  & Benoît Sagot  & 60,004/490,369\\
  & Slavic &  \texttt{pol}  &  Polish  &  \citet{wol:etal:20,wol:kier:lrec16} & Witold Kieraś, Marcin Woliński  & 274,550/13,882,543\\
  & Slavic & \texttt{ces} & Czech & \citet{morfflexcz} & Witold Kieraś & 824,074/50,284,287 \\
   \midrule
 Niger-Congo &Bantoid    &\texttt{swc}   & Swahili &\citet{swahili}   &  Jennifer White &  97/4,949\\  
 &Bantoid    &\texttt{zul}   &Zulu      & \citet{zulu}      & ---  &  87/500\\
   \midrule
Turkic & Turkic &  \texttt{tur}   & Turkish  & UniMorph \citep[Wiktionary]{kirov-etal-2018-unimorph} & Omer Goldman and Duygu Ataman  & 3,579/570,420\\
  & Turkic &  \texttt{kaz}  & Kazakh  & \cite{kazakh,kazakh2}, Polish Wiktionary & Eleanor Chodroff, Khuyagbaatar Batsuren & 1,755/40,283 \\
 \midrule
 Uralic  & Finnic &  \texttt{krl}  &  Karelian  & \citet[VepKar]{Boyko2021VepKar} & \multirow{4}{*}{\shortstack[l]{Andrew Krizhanovsky~\\ Natalia Krizhanovsky \\ Elizabeth Salesky}}  & 10,842/411,271 \\ 
 & Finnic &  \texttt{lud}  &  Ludic  &\citet[VepKar]{Boyko2021VepKar} &   & 6,751/11,313 \\
 & Finnic &  \texttt{olo}  &  Livvi  & \citet[VepKar]{Boyko2021VepKar} &  & 27,676/1,199,149 \\
 & Finnic &  \texttt{vep}  &  Veps  & \citet[VepKar]{Boyko2021VepKar} &  & 18,618/815,676 \\
 \midrule
 Kartvelian & Karto-Zan & \texttt{kat} & Georgian & \citet{guriel-etal-2022-morphological} & David Guriel & 118/21,055 \\
 \bottomrule
\end{tabular}
\end{adjustbox}
\caption{Inflectional paradigms: augmented languages. }
\label{tab:existing-lang-descr}
\end{table*}

\begin{figure*}[ht]
\begin{center}
\includegraphics[width=0.9\textwidth]{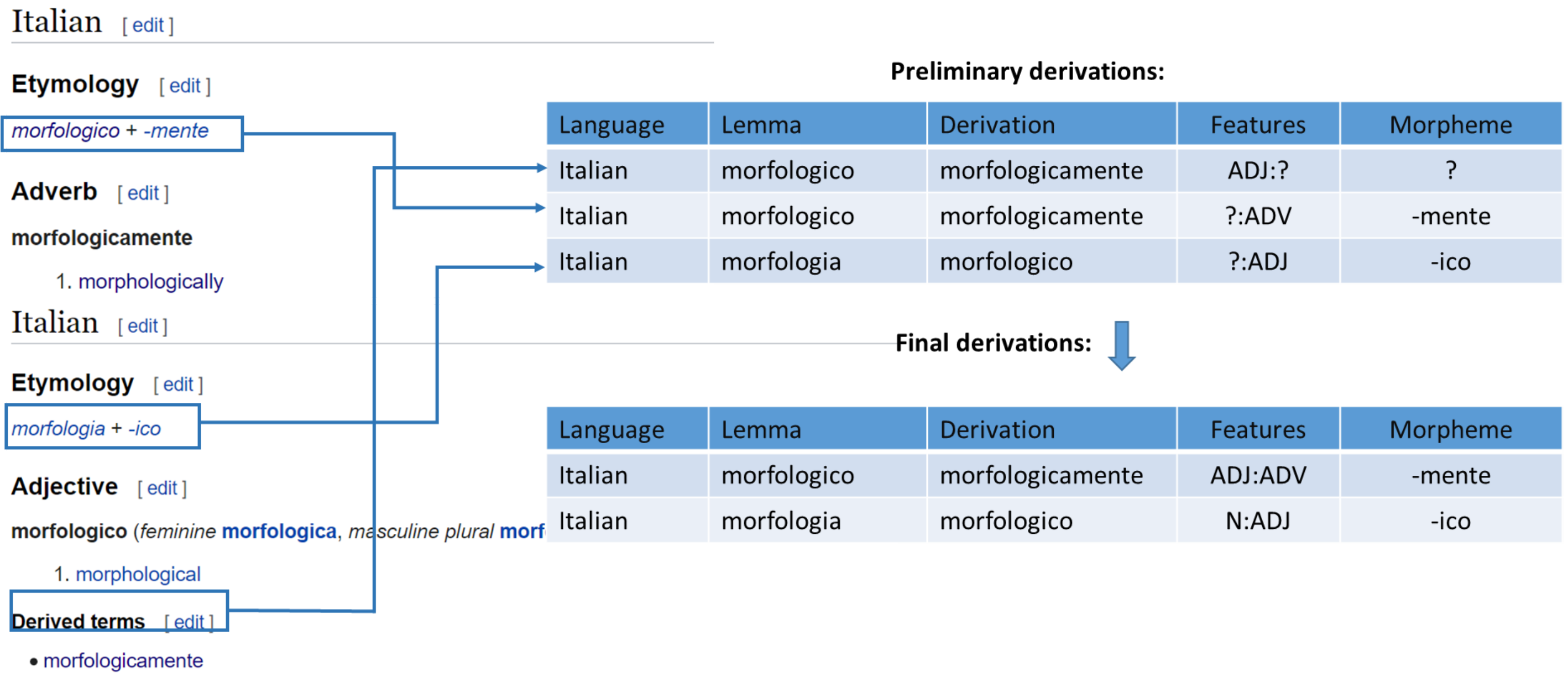}
\caption{The Wiktionary extraction process of derivational paradigms}
\label{fig:derivation}
\end{center}
\end{figure*}

\subsection{Paradigm Classes in Russian}
Aiming to establish a more granular performance analysis of (re)inflection models, we developed an application that infers possible inflection classes for each lemma present in UniMorph. By using this application, one may annotate each lemma with a set of known inflection paradigms that match all inflection samples present for a given lemma. To use this a technique, one needs a list of possible paradigms to be considered.

As a case study, we extracted a list of known inflection paradigms for Russian from the Russian edition of Wiktionary.\footnote{https://ru.wiktionary.org/} The resource provides tables of patterns which represent declension and conjugation classes as they were defined by \citet{zaliznyak2003grammaticheskij}. We merged imported patterns into a list of records each represented as a triple consisting of the following: 
\begin{itemize}[nosep]
  \item paradigm identifier (formed from a respective paradigm name given in Wiktionary);
  \item relevant \unimorph{} grammatical tags in their canonical order;
  \item word form pattern which usually contains one or more variable parts shared to other grammatical forms within the same paradigm.
\end{itemize}
We also developed an application that finds matching paradigms for every lemma in the \unimorph{} database by finding the intersection of matching paradigms over all \{lemma, form, features\} triplets observed for each given lemma in a \unimorph{} data file. Normally, multiple inflected forms occur for each lemma, which enables finding precise paradigms for most lemmas. Nevertheless, some ambiguity remains in many cases in Russian, due to the existence of numerous subtle variants in similar paradigms.

\section{New Languages and Data}
\subsection{Inflectional Paradigms}
For the UniMorph 4.0 milestone, we have added new languages scraped from linguistic resources such as Surrey Morphology Group databases \citep{azg2015}, Apertium morphological analysers \cite{tyers2010free}, and other language grammars. The current release of inflectional paradigms cover about 122 million inflections in 182 languages in total. 

\subsubsection{New Languages}
In the UniMorph 4.0 release, we introduce 67 new languages from 22 families: Afro-Asiatic, Algic, Arawakan, Austronesian, Aymaran, Chukotko-Kamchatkan, Gunwinyguan, Indo-European, Iroquoian, Koreanic, Mongolic, Niger--Congo, Oto-Manguean, Pano-Tacana, Siouan, Songhay, Trans-New Guinea, Tungusic, Turkic, Uralic, Uto-Aztecan, and Yeniseian, and the Esperanto constructed language, as shown in Table \ref{tab:new-lang-descr} and \ref{tab:new-lang-descr-2}. Of these new languages, 30 are endangered.\footnote{http://www.unesco.org/languages-atlas/index.php} Extended details on some of the languages can be found in Appendix \ref{sec:lang_details}.

\subsubsection{Augmented Languages}
The data for a handful of existing languages was expanded in several dimensions. In most cases the expansion included additions of new inflection tables from various sources, but for some languages data was added by expanding existing inflection tables (e.g.\ Turkish), by adding for more dialects (e.g.\ Arabic), or by accounting for orthographic variations (e.g.\ Hebrew). See Table \ref{tab:existing-lang-descr} for details.

For some languages the additional data is much larger. For example, the new Czech data consists of about 50M analyzed word forms from \newcite{morfflexcz}, compared to the 135k existing forms, and some Uralic languages' data grew from a few hundred forms to about a million using the VepKar corpus \citep{Boyko2021VepKar}.

\subsection{Derivational Paradigms}
\label{section:derivation}
Language-specific editions of Wiktionary contain large amounts of derivational data, typically in two forms: \emph{etymology templates} and \emph{derived terms} (see \autoref{fig:derivation}). Building on prior results from the \emph{MorphyNet} project \citep{batsuren-etal-2021-morphynet}, we have implemented an extraction mechanism from both kinds of sections, covering 12~Wiktionary editions and 30 languages. 

We managed to extract 4.3~million preliminary derivations, as reported in Table \ref{tab:derived_scrap}.
We considered such derivations as `preliminary' because they are both redundant and incomplete: some derivations are provided multiple times, but may lack indications for certain derivational features, such as parts of speech or affixes, as shown in \autoref{fig:derivation}. For example, the etymology section of the Italian \emph{`morfologia\,$\rightarrow$\,morfologico'} does not provide the part of speech of the source lemma, while \emph{`morfologico\,$\rightarrow$\,morfologicamente'} is provided in two different ways.

In order to obtain final and complete derivations, we automatically fused the preliminary instances and eliminated duplicates. As a final result, shown in Table~\ref{tab:derivations}, we inferred 769,102~derivations and 12,420~affixes for 30~languages of 10 genera.

\begin{table}[t]
\newcolumntype{R}{>{\raggedleft\arraybackslash}X}
\newcolumntype{C}{>{\centering\arraybackslash}X}
\newcolumntype{L}{>{\raggedright\arraybackslash}X}
\small 
\begin{tabularx}{\linewidth}{@{}lRr@{}}
\toprule
Wiktionary edition & Etymology & Derived terms \\
\midrule
English            & 683,351   & 1,116,122     \\
French             & 17,784    & 475,843       \\
Finnish            & 16,727    & 23,516        \\
Hungarian          & 9,358     & n.a             \\
Polish             & n.a         & 1,200,228     \\
Russian            & n.a         & 303,052       \\
German             & n.a         & 244,032       \\
Czech              & n.a         & 178,383       \\
Italian            & n.a         & 40,020        \\
Portuguese         & n.a         & 12,667        \\
Catalan            & n.a         & 7,069         \\
Serbo-Croatian     & n.a         & 4,271         \\
\addlinespace
Total              & 727,220   & 3,605,203    \\
\bottomrule
\end{tabularx}
\caption{Preliminary incomplete derivations extracted from 12 editions of Wiktionary}\label{tab:derived_scrap}
\end{table}

\begin{table}
\small
\begin{tabular}{@{}lrrr@{}}
\toprule
Languages      & Lemmas  & Derivations & Morphemes \\
\midrule
English         & 67,412 & 225,131 & 2,445 \\
Russian         & 11,922 & 93,039  & 575   \\
French          & 12,473 & 72,952  & 636   \\
Italian         & 18,650 & 58,848  & 749   \\
Polish          & 6,518  & 58,711  & 405   \\
Finnish         & 18,142 & 36,843  & 446   \\
Czech           & 4,875  & 32,336  & 318   \\
German          & 8,070  & 29,381  & 465   \\
Hungarian       & 14,566 & 28,177  & 832   \\
Spanish         & 9,159  & 25,080  & 490   \\
Dutch           & 7,810  & 13,506  & 366   \\
Portuguese      & 6,076  & 11,774  & 387   \\
Romanian        & 6,929  & 11,039  & 382   \\
Swedish         & 2,190  & 9,244   & 217   \\
Serbo-Croatian  & 4,916  & 8,553   & 429   \\
Catalan         & 5,492  & 8,284   & 241   \\
Ukraine         & 5,212  & 6,650   & 105   \\
Irish           & 3,719  & 6,417   & 270   \\
Latin           & 3,429  & 5,889   & 689   \\
Latvian         & 1,869  & 4,235   & 91    \\
Bokmal          & 2,310  & 3,238   & 227   \\
Danish          & 2,137  & 3,021   & 184   \\
Galician        & 1,995  & 2,832   & 230   \\
Greek           & 1,842  & 2,575   & 372   \\
Nynorsk         & 1,542  & 2,131   & 217   \\
Armenian        & 1,527  & 2,009   & 130   \\
Kazakh          & 1,348  & 1,965   & 91    \\
Scottish-Gaelic & 1,346  & 1,837   & 80    \\
Turkish         & 1,248  & 1,776   & 122   \\
Mongolian       & 1,410  & 1,629   & 229  \\
\addlinespace
Total          & 236,134 & 769,102     & 12,420    \\\bottomrule
\end{tabular}
\caption{Final derivations of 30 languages, released in UniMorph 4.0}\label{tab:derivations}
\end{table}

\subsection{Morpheme Segmentation}
\captionsetup[figure]{skip=2pt}

The schema update of UniMorph~3.0 \cite{mccarthy-etal-2020-unimorph} introduced segmentation structure of inflected forms along with segmented morphological features, as in \autoref{fig:segment}(c). UniMorph~4.0 extends this data structure by complete morphological analysis for 16~languages. Segmentations were computed using language-specific inflectional morpheme datasets
representing the inflection network between word forms, as shown in \autoref{fig:segment}(b). Each node of this network represents a unique set of morphological features, and each directed edge represents the fact that the target form is an inflection of the source. Each row of \autoref{fig:segment}(b) corresponds to an edge of the network, with each item in the \emph{Morphemes} column implementing the inflection. For example, in Hungarian all plural dative noun \feature{N;DAT;PL} word forms are inflected from the plural nominal \feature{N;NOM;PL} forms by one of the suffixes \textit{-ak,-ek,-ok,-ök,-k}.
Such morpheme tables were created by language expert contributors for 16~languages. Using the morpheme tables, we algorithmically (recursively) segment each inflected word form in UniMorph. This method is very effective with regular inflection cases for the 16~languages considered. In order to cover irregular inflections \cite{gorman-etal-2019-weird}, we implemented custom segmentation rules for these languages. In total, 15~million segmentations were computed for 16~languages, as shown in Table~\ref{tab:segmentation}.

Related work on segmentation or extracting lexical information from Wiktionary include the Wikinflection project \citep{metheniti-neumann-2020-wikinflection}, the DBnary project \citep{serasset2015dbnary}, MorphoChallenge data \citep{Kurimo2010ProceedingsOT}, JWKTL \citep{zesch-etal-2008-extracting}, EtymDB-2.0 \citep{fourrier-sagot-2020-methodological}, and Yawipa \citep{wu-yarowsky-2020-computational,wu-yarowsky-2020-wiktionary}.

\begin{figure}[t]
\centering
\caption*{(a) UniMorph 3.0}
\begin{tabular}{@{}lll@{}}
\toprule
Lemma & Form     & Features \\
\midrule
légy  & légy      & \texttt{N;NOM;SG} \\
légy  & legyek    & \texttt{N;NOM;PL} \\
légy  & legyeknek & \texttt{N;DAT;PL} \\
\bottomrule
\end{tabular}
\medskip
\caption*{(b) Morpheme Table\\}
\smallskip
\begin{tabular}{@{}lll@{}}
\toprule
Source Form & Morphemes & Target Form \\
\midrule
\texttt{N;NOM;SG}    & -ök;-ok;-ek;-ak;-k & \texttt{N;NOM;PL}   \\
\texttt{N;NOM;PL}    & -nak;-nek          & \texttt{N;DAT;PL}   \\
\bottomrule
\end{tabular}
\medskip
\caption*{ (c) UniMorph 4.0 with Segmentation}
\smallskip
\newcolumntype{R}{>{\raggedleft\arraybackslash}X}
\newcolumntype{C}{>{\centering\arraybackslash}X}
\newcolumntype{L}{>{\raggedright\arraybackslash}X}
\begin{tabularx}{\linewidth}{@{}llll@{}}
\toprule
Lemma & Form     & Features & Segmentation \\
\midrule
légy  & légy      & \texttt{N;NOM;SG} & ---            \\
légy  & legyek    & \texttt{N|NOM;PL} & légy\texttt{|}ek      \\
légy  & legyeknek & \texttt{N|PL|DAT} & légy\texttt{|}ek\texttt{|}nek \\
\bottomrule
\end{tabularx}
\medskip
\caption{Segmentation process}\label{fig:segment}
\end{figure}
\captionsetup[figure]{skip=5pt}

\begin{table}[]
\newcolumntype{R}{>{\raggedleft\arraybackslash}X}
\newcolumntype{C}{>{\centering\arraybackslash}X}
\newcolumntype{L}{>{\raggedright\arraybackslash}X}
\small 
\begin{tabularx}{\linewidth}{@{}lrR@{}}
\toprule
  Language     &   Lemmas   & Forms/Segmentations \\
\midrule
Finnish        & 81,729    & 3,708,296     \\
Serbo-Croatian & 68,757    & 1,760,095     \\
Latin          & 50,949    & 1,440,506     \\
Russian        & 36,387    & 1,321,024     \\
Spanish        & 65,565    & 1,289,324     \\
Hungarian      & 38,067    & 1,016,819     \\
Czech          & 33,348    & 816,956       \\
Italian        & 89,763    & 712,021       \\
Polish         & 36,940    & 663,545       \\
English        & 396,772   & 649,594       \\
German         & 39,275    & 490,331       \\
French         & 52,711    & 453,229       \\
Portuguese     & 39,029    & 376,341       \\
Catalan        & 14,979    & 158,922       \\
Swedish        & 12,508    & 131,599       \\
Mongolian      & 2,085     & 14,592        \\
\addlinespace
Total          & 1,058,864 & 15,003,194     \\\bottomrule
\end{tabularx}
\caption{UniMorph 4.0 languages with segmentations}\label{tab:segmentation}
\end{table}

\section{Validation tool}
Evaluation of morphological databases' quality is a challenging task due to the weird and irregular morphological aspects of languages \citep{gorman-etal-2019-weird}. Given millions of inflections in languages such as Finnish and Russian, manual evaluation is often time-consuming and cost-inefficient. In this release, we extend an existing UniMorph validation tool\footnote{https://github.com/unimorph/ud-compatibility}, developed by \citet{McCarthy_2018}. With this extension, we can compute the precision, recall, and F-measure for all part-of-speech categories of UniMorph resources. 
It complements the tools released in \citet{mccarthy-etal-2020-unimorph} for canonicalization and flagging common annotation errors.

With this validation tool, we evaluated five high-resource languages---English, Latin, French, Russian, and Spanish---against the UD treebanks \citep{silveira2014gold,haug2008creating,guillaume2019conversion,lyashevskaya2016universal,taule2008ancora} (Table \ref{tab:validation}). UniMorph 3.0 data results in high precision between 97.2\% and 99.8\% but at low recall rates from 10.8\% to 43.3\%. An important reason for these low recall rates was that UniMorph 3.0 was based on the data extracted 4--5 years ago. Since then, Wiktionary has been constantly improved by the Wiktionarians. Another crucial reason was the fact that UniMorph 3.0 had no inflections for adjectives and nouns for English, French, and Spanish. In addition, Latin inflections lack the entire class of deponent verbs and Russian inflections miss lexical features, e.g., gender for nouns and perfective/imperfective aspects for verbs. In both Latin and Russian, participles have no morphological features on case, gender, and number. By incorporating these into the extraction pipeline, we extracted new data from Wiktionary on these five languages and conducted the evaluation again. As shown in Table \ref{tab:validation}, recall rates were significantly improved to 61.5--89.7\% while maintaining high quality at 95.2--99.3\%.  With this approach, we have so far extended and improved 17 existing languages of UniMorph.

\begin{table}[t]
\newcolumntype{R}{>{\raggedleft\arraybackslash}X}
\newcolumntype{C}{>{\centering\arraybackslash}X}
\newcolumntype{L}{>{\raggedright\arraybackslash}X}
\small
\centering
\begin{tabular}{@{}l c c c c@{}}
\toprule
Language                 & UniMorph & Recall   & Precision    & F\textsubscript{1}   \\
\midrule
English & v3.0     & 24.6 & 98.6 & 39.4 \\
                         & v4.0     & 71.6 & 99.7 & 83.4 \\
Latin   & v3.0     & 43.3    & 97.2    & 59.9    \\
                         & v4.0     & 76.3    & 98.1    & 85.3   \\
French  & v3.0     & 20.6 & 98.5 & 34.1 \\
                         & v4.0     & 79.7 & 97.9 & 87.9 \\
Russian & v3.0     & 10.8 & 97.4 & 19.4 \\
                         & v4.0     & 61.5 & 95.2 & 74.7 \\
Spanish & v3.0     & 32.1 & 99.8   & 48.6    \\
                         & v4.0     & 89.7 & 99.3   & 94.3   \\
                         \bottomrule
\end{tabular}
\caption{Automatic validation of UniMorph v3.0 and v4.0 on UD Treebanks for five languages} \label{tab:validation}
\end{table} 



\section{Conclusion}
The UniMorph project represents a massively multilingual effort at cataloguing the world's inflectional and derivational morphology. Here, we present UniMorph 4.0 which has several improvements and expansions both in terms of contents and scopes over the previous release. First, a large community of linguists from all over the world contributed to the UniMorph project over the last few years, resulting in 67 new languages (including 30 endangered languages) and an extension of inflectional data on existing 31 languages.
Second, we amended the schema with a hierarchical structure necessary for morphological phenomena like multiple-argument agreement and case stacking, while adding missing morphological features to make the schema more inclusive.
Third, we introduced morpheme-annotated derivational paradigms, covering 769K derivations in 30 languages from 10 genera.
Fourth, we added morpheme segmentation for 16 languages. 
Finally, we implemented an automatic validation tool to evaluate the UniMorph data against the Universal Dependencies treebanks.
With all these efforts, the new release becomes more accurate and complete.
The data and tools are published under an open source license at \texttt{unimorph.github.io}.
The project  welcomes continued contributions from the community. 

\section*{Acknowledgments}

OG and RT wish to thank ERC grant no. 677352.


\section*{References}
\bibliographystyle{lrec2022-bib}
\bibliography{acl2020,acl2021,arabic,Indo-Euro,uralic,turkic,anthology,lrec2022,lrec-2022a}


\appendix

\section{Languages Details}
\label{sec:lang_details}
\subsection*{Semitic}
\paragraph{Arabic}
Modern Standard Arabic (MSA,~ara) is the primarily written form of Arabic and is used in all official communication means. 
In contrast, Arabic dialects are the primarily spoken varieties of Arabic, and the increasingly written varieties on unofficial social media platforms.
Dialects have no official status despite being widely used. Both MSA and the dialects coexist in a sate of diglossia \cite{Ferguson:1959:diglossia} whether in spoken or written form.
Arabic dialects vary among themselves and are different from MSA in most linguistic aspects (phonology, morphology, syntax, and lexical choice).
 In this work we provide inflection tables for (MSA,~ara), Egyptian Arabic (EGY,~arz), and Gulf Arabic (GLF,~afb). Egyptian Arabic is the variety of Arabic spoken in Egypt. Gulf Arabic is referred to the dialects spoken by the indigenous populations of the member states of the Gulf Cooperation Council, especially those in regions on the Arabian Gulf. 

\paragraph{Syriac} Classical Syriac is a dialect of the Aramaic language and is attested as early as the 1st century CE. As with most Semitic languages, it displays non-concatenative morphology involving primarily tri-consonantal roots. Syriac nouns and adjectives are conventionally classified into three `states'---Emphatic, Absolute, Construct---which loosely correlate with the syntactic features of definiteness, indeterminacy and the genitive. There are over 10 verbal paradigms that combine affixation slots with inflectional templates to reflect tense (past, present, future), person (first, second, third), number (singular, plural), gender (masculine, feminine, common), mood (imperative, infinitive), voice (active, passive), and derivational form (i.e., participles). Paradigmatic rules are determined by a range of linguistic factors, such as root type or phonological properties. The data included in this set was relatively small and consisted of $1{,}217$ attested lexemes in the New Testament, which were extracted from \textit{Beth Mardutho: The Syriac Institute}'s lexical database, SEDRA.

\paragraph{Hebrew}
is a member of the Northwest Semitic branch, and, like Syriac and Arabic, it is written using an abjad where the vowels are sparsely marked in unvocalized text. This fact entails that in unvocalized data the complex ablaut-extensive non-concatenative Semitic morphology is somewhat watered down as the consonants of the root frequently appear consecutively with the alternating vowel unwritten.
In this release we added data in vocalized Hebrew, in order to examine the models' ability to handle Hebrew's full-fledged Semitic morphological system.

The inflection tables are largely identical to those included in UniMorph 3.0, scraped from Wiktionary, with the addition of the verbal nouns and all forms being automatically vocalized.

\paragraph{Amharic} is the most spoken among the roughly $15$ languages in the Ethio-Semitic branch of South Semitic.
Unlike most other Semitic languages, it is written in the Ge'ez (Ethiopic) script, an abugida in which each character represents either a consonant-vowel sequence or a consonant in the syllable coda position.
Like other Semitic languages, Amharic displays both affixation and non-concatenative template morphology.
Verbs inflect for subject person, gender, and number and tense/aspect/mood.
Voice and valence are also marked, but these are treated as separate lemmas in the data.
Other verb affixes, which are not included in the data, indicate object person, gender, and number; negation; and relativization.
Nouns and adjectives share most of their morphology and are often not clearly distinguished.
Nouns and adjectives inflect for definiteness, number, and possession.
Nouns and adjectives also have prepositional prefixes and accusative suffixes, which are not included in the data.

\subsection*{Turkic}

\paragraph{Turkish} is part of the Oghuz branch, and it is highly agglutinative, like the other languages of this family.

This release vastly expanded the pre-existing UniMorph inflection tables. As with the Siberian Turkic languages, it was necessary to omit many forms from the paradigm as the UniMorph schema is not well-suited for Turkic languages. For this reason, we only included the forms that may appear in main clauses. Other than this limitation, we tried to include all possible tense-aspect-mood combinations, resulting in $30$ series of forms, each including $3$ persons and $2$ numbers. The nominal coverage is less comprehensive and includes forms with case and possessive suffixes.

\subsection*{Indo-European}
 The Indo-European language family consists most of European and Asian languages. South Asia that encompasses India, Pakistan, Bangladesh, Nepal, Bhutan, Sri Lanka and Maldives is referred to as the heartland of Indo-Aryan or Indic languages are spoken \cite{Jain:2007}. We enrich the data with two languages Magahi and Braj from Indo-Aryan or Indic languages which are spoken in Indian states. 

\paragraph{Indo-Aryan: Braj bhasha, or Braj} 
is spoken in the Western Indian states of Uttar Pradesh, Rajasthan and Madhya Pradesh, which is one of the Indo-Aryan languages. Braj is highly inflectional language in this language family. We have used the data from the literary domain \cite{Kumar:2018}. The final dataset contains 1,821 wordforms and 1,246 lexemes including nouns, verbs and adjectives. our analysis of the language has shown that there are 34 possible forms for verbs, 3 forms for adjectives and 2 forms for nouns. As is clear from this, in the first phase, we have preferred breadth (i.e.\ represent larger number of lexemes) over depth (i.e.\ only a few wordforms of most of the lexemes are represented) in the current version. 

\paragraph{Indo-Aryan: Magahi}
comes under the Magadhi group of the middle Indo-Aryan language which is spoken mainly in Eastern Indian states of Bihar and Jharkhand and also to the adjoining region of Bengal and Odisha \cite{Grierson:1903-1928}. Magahi has no grammatical gender agreement, though animate nouns like /laika/ (boy) and /laiki/ (girl) show sex-related gender derivation, noun also carry number marker that affects the form of case markers and postposition in certain instances \cite{lahiri:2021}. The language has a rich and diverse system of verbal morphology to show the honorific agreement, tense, aspect, person, resulting in as many as 24 distinct forms of verbs, 19 forms of aux and 4 forms of nouns. We have used a dataset from the literary domain in order to extract the inflectional paradigm of nouns and verbs. The present dataset contains 1,612 lexemes and 2,194 wordforms which includes noun, verb, adjective, conjunction, adverb etc. 

\paragraph{West Slavic: Upper Sorbian}
is a West Slavic language spoken by Sorbs in Germany in the historical province of Upper Lusatia, which is today part of Saxony. It is a minority language with about 13,000 speakers (\href{https://www.ethnologue.com/language/HSB}{Ethnologue}). The Upper Sorbian dataset contains 310 word forms and 400 lemmas. The data source is the corpus compiled by the Sorbian Institute and The Witaj Sprachzentrum in Germany, that was used as a training model for an unsupervised MT task \cite{fraser-2020-findings}. All conjugated parts of speech existing in the language are presented in the dataset. Adjectives, when plural or dual, are marked with case only, otherwise have gender marking, according to Upper-Sorbian grammar. 

\paragraph{West Slavic: Czech, Polish, Slovak}
Data for three West Slavic languages has been added or updated from sources outside Wiktionary. These are: Polish, Czech and Slovak. All three are closely related and are highly inflectional. The Polish data comes from the \emph{Grammatical Dictionary of Polish} \citep{wol:etal:20,wol:kier:lrec16}, an extensive database consisting of inflectional paradigms for Polish lexemes. It serves both as a~standalone electronic dictionary as well as a source data for morphological analysers and other applications. The dictionary allows for exporting its data in various schemes so it was possible to prepare a separate exporting path directly for the UniMorph annotation scheme. In the final data all proper names were omitted. The dataset consists of 13,882,543 wordforms of 274,550 lexemes. 

The Czech and Slovak data were obtained from the LINDAT/CLARIAH repository \citep{morfflexcz}, \citep{morfflexsk}. Both datasets were intended for the use in morphological analysers and their grammatical information is represented in the native Czech National Corpus tagset. The datasets were converted automatically to the UniMorph scheme. Proper names as well as some archaic and non-standard wordforms were omitted. Additionally to limit the size of both data collections negated forms of nouns and adjectives which are perfectly regular were also omitted. The final Czech dataset consists of 50,284,287 wordforms of 824,074 lexemes and the Slovak one contains 28,428,612 wordforms of 366,183 lexemes. 

\paragraph{East South Slavic: Pomak}
Pomak (endonym: Pomácko, Pomáhcku or other dialectic variants) is a non-standardised East South Slavic (ESS) language variety mainly spoken in the region of Greek Thrace, as well as in places of Pomak diaspora. Pomak is included in the map of the European Languages Equality Network.\footnote{\url{https://elen.ngo/languages-map/}} In comparison to all ESS languages, Pomak exhibits a more profound phonological, morphological, morphosyntactic and lexical influence by Medieval and Modern Greek and, due to the predominantly Muslim religion of its speakers, a more profound lexical and phonotactical influence by Ottoman and Modern Turkish. The Pomak data were collected by linguist and native Pomak speaker Ritván Karahóǧa, under the “PHILOTIS: State-of-the-art technologies for the recording, analysis and documentation of living languages” project (MIS 5047429), which is implemented under the “Action for the Support of Regional Excellence”, funded by the Operational Programme “Competitiveness, Entrepreneurship and Innovation” (NSRF 2014-2020) and co-financed by Greece and the European Union (European Regional Development Fund).
The final dataset includes 233,533 lemmas and a total of 6,557,759 wordforms covering adjectives, nouns, and verbs.

\subsection*{Uralic}

In 2019–2020 generation algorithms of nominal and verbal wordform were developed for the Veps language, Livvi Karelian and Karelian Proper.\footnote{See formalized morphological inflectional rules in Veps and Karelian:  \href{https://figshare.com/projects/VepKar/100664}{https://figshare.com/projects/VepKar/100664}} Due to this implementation, 2.1 million word forms were generated in the VepKar corpus in the semi-automatic mode during the last two years.

Data for Uralic languages (Karelian, Ludic, Livvi and Veps) were exported from the VepKar corpus~\citep{Boyko2021VepKar}. The VepKar dataset consists of more than 2,4 million wordforms of approximately 64 thousand lemmas.

\subsection*{Austronesian}
Austronesian languages are widely spoken throughout Taiwan, Greater Central Philippines, Madagascar, Islands of Southeast Asia, and Pacific Islands. Derivational and inflectional morphology of languages in this family rely on prefixation and suffixation; some infixation and circumfixation are also attested, as found in Tagalog and Indonesian respectively \citep{LevinPolinsky2019}. In this language family, reduplication is also common \citep{GhanggoAte2021}. In Indonesian, a morphologically rich language, prefixation, suffixation, and circumfixation function in both verb-forming and noun-forming processes. In addition, in the verbal system, main morphological exponents mark voice distinctions as well as active and passive or causatives and applicatives.  For some languages whose affixes are moderate in number, clitics are pervasive and morphological exponents mark voice distinction may be lost. Kodhi/Kodi, a language of the Sumba-Hawu sub-group, is the prime example. In this language, pronouns, emphatic, perfective aspect, politeness are expressed by attaching clitics to nouns, verbs, and adjectives. In terms of pronominal clitics, they co-occur with free pronouns marking TERM relations (subjects and objects) and possession, and function like a system of agreement. Kodhi/Kodi also shows loss of Austronesian voice morphology which is typically found in Indonesian-type languages \citep{arka2002}.

\subsection*{Iroquoian}

As a member of the Iroquoian (Hodinöhšöni)
language family, the Seneca language is an indigenous Native American language that is considered critically endangered. Currently the language is estimated to have fewer than 50 first-language speakers left and most of them are elders. The language is spoken mainly in three
reservations located in Western New York: Allegany, Cattaraugus, and Tonawanda.
Seneca has high (inflectional) morphological complexity, containing agglutinative as well as fusional properties.

\subsection*{Arawak and Pano-Takana}
We include three languages from the Amazon region:
\paragraph{Asháninka} is an Arawak language spoken along the rivers Tambo, Ene, Apurímac, Urubamba y Bajo Perené in Central Peruvian Amazon. It belongs to the Ash\'aninka-Ash\'eninka dialect complex, which comprises more than 70,000 speakers in Central and Eastern Peru and in the state of Acre in Eastern Brazil \cite{pedros_2018}. Ash\'aninka belongs to the Nihagantsi subgroup, previously known as Campa in the literature.  
Asháninka is an agglutinating, polysynthetic, verb-initial language. Since it is a strongly head-marking language, the verb is the most morphologically complex word class, with a rich repertoire of aspectual and modal categories. The language lacks case marking, except for one locative suffix; grammatical relations of subject and object are indexed as affixes on the verb itself. 
The corpus consists of inflected nouns and verbs from the variety spoken in the Tambo river of Central Peru. The annotated nouns take possessor prefixes, locative case and/or plural marking, while the annotated verbs take subject prefixes, reality status (realis/irrealis), and/or perfective aspect. 
\paragraph{Yanesha'} is an Arawak language from the Pre-Andine branch. It is spoken in Central Peru by between 3,000 - 5,000 people. 
Yanesha' is an agglutinating, polysynthetic language with a VSO constituent order. Nouns and verbs are the two major parts of speech. The existence of an independent class of adjectives is questionable due to the absence of clear non-derived forms. Yanesha' is strongly head-marking and therefore the verb class is the most morphologically complex lexical class and the only obligatory constituent of a clause. \cite{dixon_aikhenvald_2001}. 
The corpus consists of inflected nouns and verbs from both dialectal varieties. The annotated nouns take possessor prefixes, plural marking, and locative case, while the annotated verbs take subject prefixes.

\paragraph{Shipibo-Konibo} is a Panoan language spoken by around 35,000 native speakers in the Amazon region of Peru. Its morphology is mainly agglutinating, synthetic and almost exclusively suffixing (with only a closed set of prefix related to body-part concepts) Word order is pragmatically determined, but there is some tendency towards SOV constructions. Verbs lack subject and object markers, but exhibit a relatively complex set of TAME markers. As with other Panoan language, verbs is Shipibo-Konibo are strictly transitive or intransitive, with almost no cases of labile verbs in the language. Other relevant grammatical categories for Shipibo-Konibo are participant agreement, switch reference and evidentiality. Data for Shipibo-Konibo were extracted mainly from an old dictionary \cite{james1993diccionario} and a grammar \cite{valenzuela2003transitivity}.

\subsection*{Koreanic}
\paragraph{Korean} is an East Asian isolate language spoken by about 80 million people. The dataset was compiled using Wiktionary inflection tables. The resulting data is 2,686 lemmas and 241,323 word forms. It consists of mostly predicates, so the resulting lemmas are mainly verbs and a smaller number of adjectives. The scraped annotated paradigms turned out to be quite similar (mainly because the adjective paradigm is a reduced verb paradigm) and do not represent all forms of verbs and adjectives. It is important to note that different types of converbs were tagged consistently. 

\subsection*{Yeniseian}
\paragraph{Ket} is the only surviving language of the Yeniseian family with about 60 speakers of all levels of linguistic competence (\href{https://minlang.iling-ran.ru/lang/ketskiy}{Minlang}). The data source is a text collection compiled during the field work of the Laboratory for Computational Lexicography of the Moscow State University, that took place between 2004 and 2009. The Ket dataset contains the word forms of 12 categories, 7 of them (ADJ, NUM, ADV, INTJ, ADP, PART, CONJ) are invariable. The complexity of the Ket verb consists in polypersonal conjugation. The case and number of all arguments object and subject are reflected  in the verb.

\end{document}